\documentclass[runningheads]{llncs}

\usepackage{eccv}

\usepackage{eccvabbrv}
\usepackage[accsupp]{axessibility}
\usepackage{graphics} 
\usepackage{graphicx} 
\usepackage{amsmath} 
\usepackage{amssymb}  
\usepackage[pagebackref,breaklinks,colorlinks,citecolor=eccvblue]{hyperref}
\usepackage{multirow}
\usepackage{booktabs}
\usepackage{siunitx}
\usepackage{tikz}
\usepackage{orcidlink}
\usepackage[dvipsnames]{xcolor}

\usetikzlibrary{shapes, arrows, positioning, fit, calc, backgrounds, shadows}
\usepackage{amsmath}
\begin{document}

\title{(MGS)$^2$-Net: Unifying Micro-Geometric Scale and Macro-Geometric Structure for Cross-View Geo-Localization} 

\titlerunning{(MGS)$^2$-Net}

\author{Minglei Li\inst{1}\orcidlink{0009-0007-3454-8073} \and
Mengfan He\inst{1} \and
Chunyu Li\inst{1} \and
Chao Chen\inst{1} \and
Xingyu Shao\inst{1} \and
Ziyang Meng\inst{1}}

\authorrunning{M. Li et al.}

\institute{Department of Precision Instrument, Tsinghua University, Beijing 100084, China
\email{liminglei25@mails.tsinghua.edu.cn}\\
}

\maketitle

\begin{abstract}
In this paper, we propose (MGS)$^2$-Net, a geometry-grounded framework that introduces an effective Macro-Geometric Structure Filtering (MGS-F) module to cross-view geo-localization (CVGL). Unlike pixel-wise matching that is commonly used, MGS-F leverages dilated geometric gradients to physically filter out high-frequency facade interferences while emphasizing the view-invariant features on the horizontal plane, directly addressing the oblique-orthogonal view difference. To address the severe scale variations caused by varying UAV flight altitudes, we explicitly incorporate another Micro-Geometric Scale Adaptation (MGS-A) module. MGS-A dynamically utilizes depth priors to rectify scale discrepancies via multi-branch feature fusion. Furthermore, a Structure-Guided Contrastive (SGC) loss is designed to strictly discriminate against cross-view blind spots. Extensive experiments demonstrate that (MGS)$^2$-Net achieves state-of-the-art performance, achieving a Recall@1 of 97.60\% on the University-1652 dataset and 98.45\% on the SUES-200 dataset. Furthermore, the framework exhibits superior cross-dataset generalization against geometric ambiguity.
\keywords{Cross-View Geo-Localization \and Unmanned Aerial Vehicles \and Geometric Filtering  \and Scale Adaptation}

\end{abstract}


\section{Introduction}
\vspace{-5pt}
Unmanned Aerial Vehicles (UAVs) have broad applications from urban monitoring to autonomous delivery \cite{1}, \cite{2}, \cite{55}, and achieving robust localization is critical for autonomous flight of UAVs. Since Global Navigation Satellite Systems (GNSS) are often degraded due to signal jamming and multipath effects in complex urban environments \cite{3}, Cross-View Geo-Localization (CVGL), matching real-time onboard views against satellite orthophotos, has emerged as an alternative solution \cite{40} \cite{42}.

CVGL is challenged by extreme domain shifts: UAV views contain rich multi-perspective imagery, while satellite views are strictly orthographic \cite{4}. To bridge this gap, the current trend is to transfer from CNN-based metric learning \cite{5}, \cite{41} to Vision Transformer (ViT) approaches like TransGeo \cite{6}, and L2LTR \cite{16}. Recent works have further refined feature granularity through content-aware hierarchical representation selection \cite{57} and correlation-aware learning \cite{58}.

Despite these advancements, existing methods lack explicit 3D structural awareness. In dense urban scenarios, the primary bottleneck is macro-level structural noise: oblique UAV images feature a greater number of vertical facades that serve as high-frequency noise that is not available in satellite maps. This structural gap is further complicated by micro-level scale ambiguity, where texture sizes vary drastically with depth, making feature alignment unreliable. Merely concatenating depth maps is insufficient, as raw monocular depth lacks the semantic capacity to distinguish valid matching surfaces from invalid ones.

We propose (MGS)$^2$-Net, a novel network that shifts the CVGL solution from passive texture matching to active geometric filtering. The insight here is that robust alignment hinges on explicitly filtering out view-dependent vertical structures. To this end, we introduce a Macro-Geometric Structure Filtering (MGS-F) module to serve as the backbone of our framework. MGS-F computes dilated geometric gradients to capture large-scale planar trends, effectively suppressing vertical facade features while preserving horizontal plane structure features.

\begin{figure}[!t]
  \centering
  \begin{minipage}[t]{0.48\columnwidth}
    \centering
    \includegraphics[height=6cm]{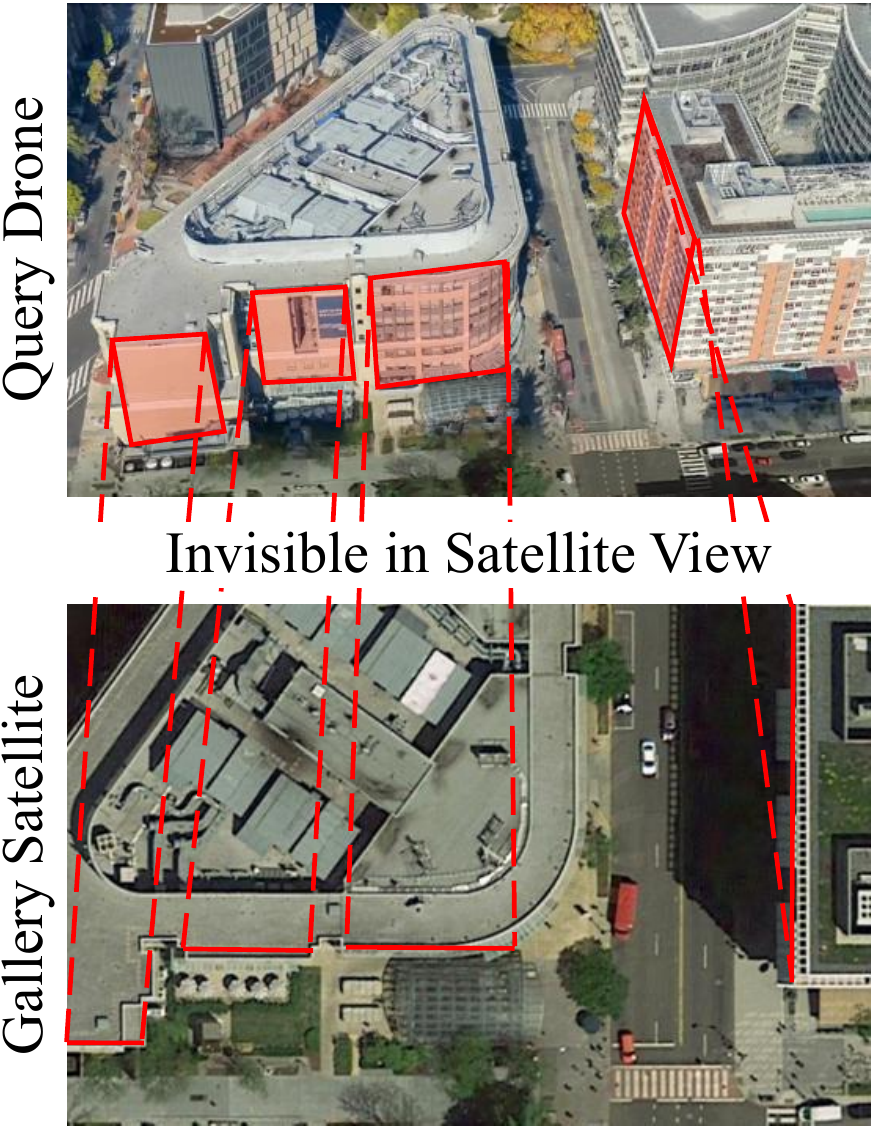}
    \centerline{\footnotesize(a) Appearance Ambiguity (Baseline)}
  \end{minipage}%
  \begin{minipage}[t]{0.48\columnwidth}
    \centering
    \includegraphics[height=6cm]{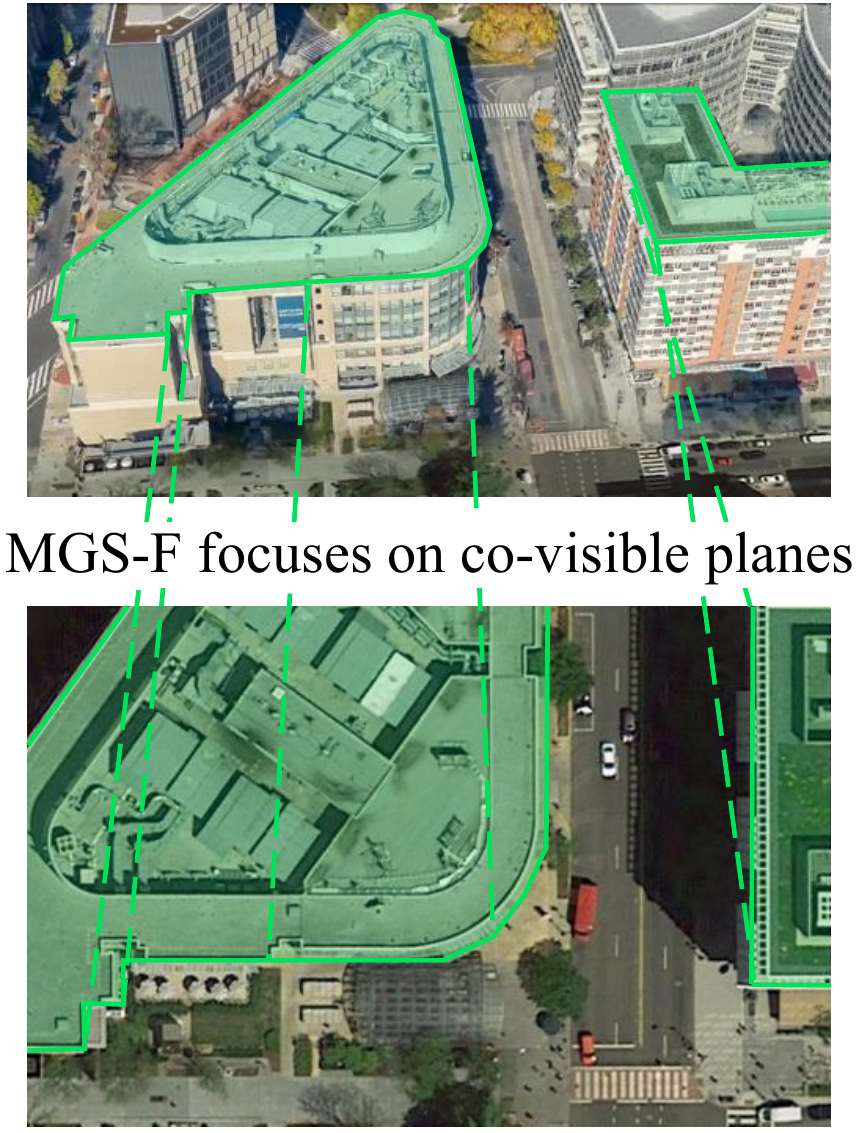}
    \centerline{\footnotesize(b) Geometric Filtering (Ours)}
  \end{minipage}
  \caption{From Texture Dependency to Geometric Grounding. (a) Visual Ambiguity: Existing methods often overfit to view-dependent vertical facades (\textcolor{red}{red} boxes) that are invisible in the satellite orthophoto, leading to retrieval failure. (b) Our Solution: MGS-Net filters out these vertical interferences via MGS-F and robustly focuses on view-invariant rooftops (\textcolor{ForestGreen}{green} boxes), ensuring consistent cross-view alignment.}
  \label{fig1}
\end{figure}

Complementing MGS-F, we introduce the Micro-Geometric Scale Adaptation (MGS-A) module to resolve the scale ambiguity that often undermines structural analysis. By applying depth priors and regressing dynamic attention weights, MGS-A fuses features from three different scale-expanding branches. Finally, the Structure-Guided Contrastive (SGC) Loss discriminates against cross-view blind spots by enforcing a margin between horizontal and vertical features.

Our contributions can be summarized as follows:
\begin{itemize}
\item We propose a geometry-grounded CVGL paradigm via (MGS)$^2$-Net. To the best of our knowledge, this is the first framework to explicitly leverage 3D structural constraints to bridge the oblique-orthogonal view difference. By introducing the MGS-F module that filters out view-dependent vertical interferences, we successfully shift the considered problem from passive 2D texture matching to active 3D geometric alignment.
\item We propose a synergistic combination of the MGS-A module and the SGC Loss to further improve geometric consistency. The MGS-A module utilizes depth priors to dynamically rectify scale ambiguity across varying flight altitudes, while the SGC Loss treats vertical facades as hard negatives, constraining the network to differentiate cross-view blind spots.
\item We achieve new state-of-the-art performance across multiple benchmarks. Extensive experiments demonstrate that our method delivers record-breaking Recall@1 scores of 97.60\% on the University-1652 dataset and 98.45\% on the SUES-200 dataset. Furthermore, (MGS)$^2$-Net exhibits strong cross-dataset generalization capabilities, verifying the superior robustness of the proposed approach against domain shifts.
\end{itemize}

\vspace{-3pt}
\section{Related Work \label{2}}
\vspace{-5pt}

\subsection{Foundation Models and Cross-View Geo-Localization} 
\vspace{-2pt}
Cross-view geo-localization (CVGL) is normally formulated as a metric learning problem, aiming to embed location-dependent visual cues into a shared latent space \cite{11}. Early works predominantly relied on Convolutional Neural Networks (CNNs). Approaches like CVM-Net \cite{13} utilized NetVLAD \cite{64} for global descriptor aggregation, while RK-Net \cite{14} incorporated keypoint detection to focus on discriminative regions.

Recently, the introduction of Vision Transformers (ViT) and large-scale Foundation Models has marked a paradigm shift \cite{56}. For instance, Sample4Geo \cite{7} achieved significant gains via hard-negative mining. Concurrently, foundation models like DINO \cite{34} and DINOv2 \cite{35} have demonstrated exceptional generalization capabilities. Recent works \cite{36}, \cite{37} have successfully adapted these pre-trained frozen backbones to UAV problem, utilizing their robust semantic descriptors to mitigate domain shifts. Despite their representational power, these RGB foundation models operate on a 2D manifold. Consequently, they neglect depth-induced scale ambiguities and overfit to visual interferences rather than geometrically consistent structures.

\subsection{Fine-Grained Alignment and Semantic Fusion} 
\vspace{-2pt}
To address the spatial misalignment inherent in global descriptors, research has shifted towards fine-grained semantic alignment \cite{23}. Several approaches integrate high-level semantics through graph convolutional networks \cite{20}, correlation estimators \cite{59}, or coarse-to-fine regional alignments \cite{60} to model layout consistencies. To handle diverse scene characteristics, multi-environment adaptive frameworks like MESAN \cite{22} have been proposed. Other advancements focus on refining feature correspondence through hierarchical distillation \cite{17}, mixing features for robust visual place recognition \cite{61}, or unifying retrieval and reranking pipelines \cite{26}. However, since the current semantic methods lack explicit 3D constraints, texturally similar but geometrically distinct planes are often misaligned, severely degrading the performance of CVGL.
\vspace{-2pt}
\subsection{Geometry-Aware Cross-View Geo-Localization} 
\vspace{-2pt}
Recognizing the limitations of 2D texture matching, a growing number of studies have begun to incorporate geometric cues. Generative methods synthesize novel views to bridge the perspective gap, such as translating satellite imagery into street-level panoramas \cite{62}, while Wang et al. \cite{28} utilize Neural ODEs for continuous manifold modeling. In broader place recognition tasks, LiDAR-camera fusion \cite{29}, \cite{30} has been used to impose metric constraints. However, in standard CVGL, where only monocular images are available, existing methods almost entirely neglect the utilization of depth information \cite{31}, \cite{32}. This 2D matching paradigm is inherently suboptimal, as it lacks the 3D geometric awareness required to differentiate between view-invariant structures and view-dependent interferences. Diverging from these conventional 2D approaches, we propose a geometry-grounded paradigm via (MGS)$^2$-Net, where the state-of-the-art Depth Anything 3 \cite{12} is leveraged to extract explicit depth priors.

\vspace{-2pt}
\section{Methodology \label{3}}
\vspace{-2pt}
\begin{figure*}[!t]
\centering\includegraphics[width=12cm]{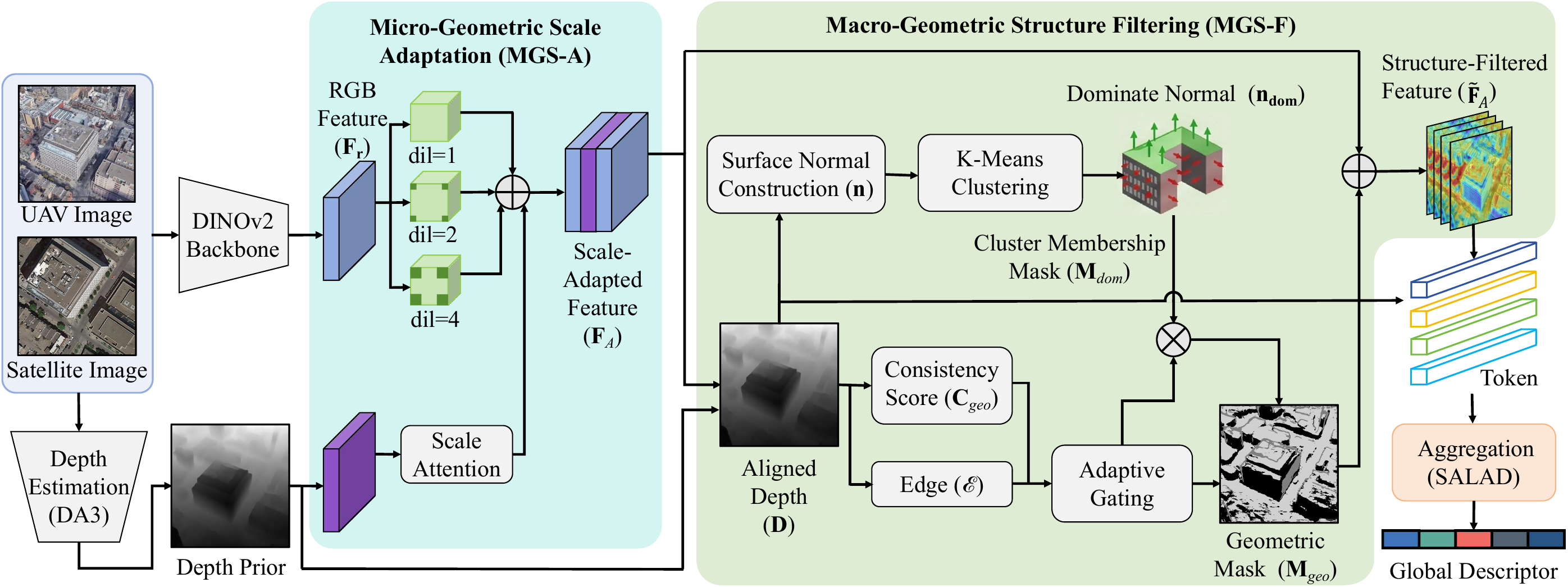}
\caption{Overview of the proposed geometry-grounded (MGS)$^2$-Net framework. The pipeline decomposes cross-view alignment into two stages: micro-level scale adaptation and macro-level structural filtering, ultimately yielding view-invariant representations.}
\label{fig2}
\end{figure*}

\subsection{Overview}
\vspace{-2pt}
Given a query UAV image $I_u$ and a gallery of satellite images $\mathcal{S} = \{I_s^i\}_{i=1}^N$, cross-view geo-localization aims to learn a mapping $\mathcal{F}(\cdot)$ that projects both domains into a shared latent space. A fundamental challenge is the domain gap: UAV images capture oblique views that contain vertical facades, whereas satellite images feature orthographic horizontal planes. To mitigate these view-dependent interferences, we propose a geometry-aware framework comprising these novel components (Figure \ref{fig2}).

\subsection{Micro-Geometric Scale Adaptation (MGS-A)}
\vspace{-2pt}
To mitigate the scale ambiguity caused by varying UAV flight altitudes, MGS-A dynamically calibrates the receptive field guided by depth cues. Given an initial RGB feature map $\mathbf{F}_r \in \mathbb{R}^{C \times H \times W}$ from the backbone \cite{35}, MGS-A generates 3 branches $\mathbf{F}_k$ for near ($k=1$), middle ($k=2$), and far ($k=3$) spatial scales:

\begin{align}
\mathbf{F}_1 = \mathbf{F}_r, \quad \mathbf{F}_2 = \phi_{2}(\mathbf{F}_r), \quad \mathbf{F}_3 = \phi_{3}(\mathbf{F}_r),
\label{eq1}
\end{align}
where $\phi_{2}$ and $\phi_{3}$ denote depth-wise $3 \times 3$ convolutional layers with dilation rates of 2 and 4, respectively.

To determine the optimal scale contribution, the normalized depth map is first encoded into depth embeddings $\mathbf{F}_d \in \mathbb{R}^{C \times H \times W}$. A predictive convolutional head $\psi$ projects $\mathbf{F}_d$ into scale-specific logits $\mathbf{Z} = \psi(\mathbf{F}_d) \in \mathbb{R}^{3 \times H \times W}$. Let $\mathbf{Z}_k \in \mathbb{R}^{H \times W}$, k = 1, 2, 3, denote the $k$-th channel slice of $\mathbf{Z}$. The dynamic weight map for the $k$-th scale branch, $\mathbf{W}_k \in [0, 1]^{H \times W}$, is then computed via a channel-wise exponential normalization:

\begin{align}
\mathbf{W}_k = \frac{\exp(\mathbf{Z}_k)}{\sum_{c=1}^{3} \exp(\mathbf{Z}_c)}.
\end{align}
The scale-adaptive feature $\mathbf{F}_{A} \in \mathbb{R}^{C \times H \times W}$ is obtained via a weighted fusion, followed by a residual connection to preserve the original semantic information:

\begin{align}
\mathbf{F}_{A} = \mathbf{F}_r + \sum_{k=1}^3 \mathbf{W}_k \odot \mathbf{F}_k,
\end{align}
where $\mathbf{W}_k$ represents the $k$-th spatial slice of the weight tensor broadcasted along the channel dimension, and $\odot$ denotes the Hadamard product. The resulting feature map $\mathbf{F}_{A}$ possesses enhanced scale invariance and serves as the refined input to the subsequent structure filtering module.
\vspace{-2pt}
\subsection{Macro-Geometric Structure Filtering (MGS-F)}
\vspace{-2pt}
MGS-F extracts a dominant geometric normal as a robust prior to indicate the co-visible planar direction, effectively suppressing cross-view blind spots.

\subsubsection{Macro Gradient and Edge Perception.}
Given a bilinearly aligned raw depth map $\mathbf{D} \in \mathbb{R}^{H \times W}$, we compute depth gradients using dilated Sobel convolutions to capture large-scale planar orientations while suppressing local roughness:
\begin{align}
\mathbf{G}_x = \mathbf{D} * \mathbf{S}_x^{(r)}, \quad \mathbf{G}_y = \mathbf{D} * \mathbf{S}_y^{(r)},
\end{align}
where $*$ denotes the convolution operation, and $\mathbf{S}_x^{(r)}, \mathbf{S}_y^{(r)}$ are predefined horizontal and vertical Sobel kernels with a dilation rate of $r$.

To identify and suppress discontinuity-dominated regions, we compute the gradient magnitude tensor $\mathbf{G}_{mag} = \sqrt{\mathbf{G}_x^2 + \mathbf{G}_y^2}$, where the operations are applied element-wise. High-frequency ambiguous edges are flagged using a binary mask $\mathbf{M}_{edge} = \mathbb{I}(\mathbf{G}_{mag} > \tau_{grad})$, where the logical inequality is evaluated element-wise across the spatial dimensions, $\mathbb{I}(\cdot)$ denotes the indicator function, and $\tau_{grad} \in \mathbb{R}$ is an adaptive scalar threshold.

\subsubsection{Dominant Plane Perception via Normal Clustering.}
The dense surface normal tensor $\mathbf{N} \in \mathbb{R}^{3 \times H \times W}$ is analytically constructed from the derived gradients via $L_2$ normalization along the channel dimension:

\begin{equation}
\mathbf{N} = \frac{[-\mathbf{G}_x, -\mathbf{G}_y, \mathbf{1}]}{|[-\mathbf{G}_x, -\mathbf{G}_y, \mathbf{1}]|_2},
\end{equation}
where $[\cdot , \cdot]$ denotes channel-wise concatenation and $\mathbf{1} \in \mathbb{R}^{H \times W}$ is an all-ones tensor matching the spatial dimensions.

To robustly identify the co-visible plane, we perform K-Means clustering using cosine similarity on the normal vectors strictly outside the edge regions. This partitions the spatial normals, from which we extract the centroid of the largest cluster as the dominant normal vector $\mathbf{n}_{dom} \in \mathbb{R}^{3}$. Concurrently, the spatial footprint of this cluster is captured by a binary indicator mask $\mathbf{M}_{dom} \in \{0, 1\}^{H \times W}$, where values are 1 if the corresponding normal belongs to the dominant cluster.

Subsequently, the continuous geometric consistency map $\mathbf{C}_{geo} \in \mathbb{R}^{H \times W}$ is computed by projecting the local normals onto this dominant direction:
\begin{equation}
\mathbf{C}_{geo} = \mathbf{n}_{dom}^\top \mathbf{N},
\end{equation}
where the projection is applied along the channel dimension of $\mathbf{N}$.

\subsubsection{Adaptive Gating and Residual Modulation.}
A learnable Sigmoid activation $\sigma(\cdot)$ with scale $s$ and bias $\beta$ is applied to the consistency map. To prevent destructive modulation on ambiguous boundaries, regions identified by the edge mask are strictly rectified to a learnable neutral prior $\mu$. The unified continuous geometric mask tensor $\mathbf{M}_{geo} \in [0, 1]^{H \times W}$ is formulated as:

\begin{equation}
\mathbf{M}_{geo} = (\mathbf{1} - \mathbf{M}_{edge}) \odot \Big( \sigma(s \mathbf{C}_{geo} + \beta) \odot \mathbf{M}_{dom} \Big) + \mu \mathbf{M}_{edge},
\end{equation}
where $\mu \in \mathbb{R}$ is a learnable scalar initialized to $0.5$ to represent the baseline structural confidence. This robust geometric prior is then injected into the semantic stream via additive residual modulation:

\begin{align}
\tilde{\mathbf{F}}_A = \mathbf{F}_A + \lambda \mathbf{M}_{geo},
\end{align}
where $\lambda$ is a learnable scalar balancing the geometric injection, and the spatial mask $\mathbf{M}_{geo}$ is broadcast along the channel dimension to match the shape of $\mathbf{F}_A$. This design strictly preserves the underlying semantic stability while actively highlighting the structural cues of co-visible planes.

\subsubsection{Depth-Conditioned Global Aggregation.}

Given the depth embeddings $\mathbf{F}_{d} \in \mathbb{R}^{C \times H \times W}$ previously encoded in MGS-A, a spatial attention gate is generated via a lightweight $1\times 1$ convolutional bottleneck $\phi_{spa}$ to selectively enrich the semantic features $\tilde{\mathbf{F}}_A \in \mathbb{R}^{C \times H \times W}$:
\begin{align}
\hat{\mathbf{F}}_A = \tilde{\mathbf{F}}_A \odot \big(\mathbf{1} + \phi_{spa}(\mathbf{F}_{d})\big),
\end{align}
where $\phi_{spa}(\mathbf{F}_{dep}) \in \mathbb{R}^{H \times W}$ acts as a spatial modulator and is broadcast along the channel dimension. To explicitly embed 3D spatial awareness, the 2D grid coordinates $\mathbf{P}_{grid} \in \mathbb{R}^{2 \times H \times W}$ are channel-wise concatenated with the aligned depth map $\mathbf{D} \in \mathbb{R}^{H \times W}$. A spatial bias is then injected using a neural mapping function $\varphi_{pos}$, which projects the $3 \times H \times W$ joint coordinate tensor into the semantic space:

\begin{align}
\mathbf{F}_{agg} = \hat{\mathbf{F}}_A + \gamma \varphi_{pos}([\mathbf{D}; \mathbf{P}_{grid}]),
\end{align}
where $\gamma \in \mathbb{R}$ is a learnable scalar, and $\mathbf{F}_{agg} \in \mathbb{R}^{C \times H \times W}$ represents the spatially enriched feature map. Finally, $\mathbf{F}_{agg}$ and the semantic tokens are processed through the core SALAD aggregation module \cite{54}, leveraging Sinkhorn optimal transport and a dustbin mechanism, followed by $L_2$ normalization to yield the final global descriptor.

\subsection{Structure-Guided Contrastive (SGC) Loss}
To prevent the backbone from implicitly overfitting to prominent vertical textures, we propose SGC, ensuring stronger feature activation on co-visible planes than on hard negatives.

\subsubsection{Geometric Partitioning.}
We define two binary partitioning masks for the co-visible regions ($\mathbf{M}_{\mathcal{P}}$) and non-covisible regions ($\mathbf{M}_{\mathcal{N}}$):

\begin{align}
\mathbf{M}_{\mathcal{P}} = \mathbb{I}(\mathbf{M}_{geo} > \tau_{high}), \quad \mathbf{M}_{\mathcal{N}} = \mathbb{I}(\mathbf{M}_{geo} < \tau_{low}),
\end{align}
where $\mathbb{I}(\cdot)$ denotes the indicator function, and both logical inequalities are evaluated element-wise across the spatial dimensions. The variables $\tau_{high}, \tau_{low} \in \mathbb{R}$ represent dynamically computed scalar thresholds for each image.

\subsubsection{Activation Aggregation.}
To quantify the network's spatial attention, we define the semantic activation map $\mathbf{A}_{sem} \in \mathbb{R}^{H \times W}$ as the channel-wise mean absolute magnitude of the geometrically enhanced features $\tilde{\mathbf{F}}_A$:

\begin{align}
\mathbf{A}_{sem} = \frac{1}{C} \sum_{c=1}^C |\tilde{\mathbf{F}}_{A, c}|,
\end{align}
where $c$ denotes the channel index. We then aggregate the mean activation intensities for the partitioned regions. The average activations $v_{\mathcal{P}}$ and $v_{\mathcal{N}}$ for the partitioned regions are formulated as:

\begin{align}
v_{\mathcal{P}} = \frac{\| \mathbf{A}_{sem} \odot \mathbf{M}_{\mathcal{P}} \|_1}{\| \mathbf{M}_{\mathcal{P}} \|_1}, \quad v_{\mathcal{N}} = \frac{\| \mathbf{A}_{sem} \odot \mathbf{M}_{\mathcal{N}} \|_1}{\| \mathbf{M}_{\mathcal{N}} \|_1}.
\end{align}
The scalars $v_{\mathcal{P}}$ and $v_{\mathcal{N}}$ represent the network's mean response to invariant co-visible planes and variant vertical facades, respectively.
\vspace{-3pt}
\subsubsection{Contrastive Ranking Objective.}
The SGC loss enforces a predefined margin $\xi$ to penalize higher activations on facade-like regions compared to horizontal planes. We formulate this objective using the standard hinge loss notation:

\begin{align}
\mathcal{L}_{SGC} = [v_{\mathcal{N}} - v_{\mathcal{P}} + \xi]_+,
\end{align}
where $[\cdot]_+ = \max(0, \cdot)$ denotes the standard hinge operation. As a semantic rectifier, $\mathcal{L}_{SGC}$ distills relative activation ordering into the feature representations.
\vspace{-12pt}
\subsection{Optimization Objective}
The overall training objective combines the primary place-retrieval loss with the proposed geometric constraint:

\begin{align}
\mathcal{L}_{total} = \mathcal{L}_{ret} + \lambda_{geo} \mathcal{L}_{SGC},
\end{align}
where $\mathcal{L}_{ret}$ is the baseline metric learning objective, i.e., the Multi-Similarity loss \cite{63}, and $\lambda_{geo}$ is a scaling factor balancing the geometric regularization.

\vspace{-2pt}
\section{Experiments \label{4}}
\vspace{-1pt}
\subsection{Datasets and Evaluation Metrics}
\vspace{-1pt}
To evaluate the effectiveness of the proposed framework, we conducted experiments on two standard benchmarks: University-1652 \cite{38} and SUES-200 \cite{39}.

University-1652 is a large-scale multi-view dataset containing 1,652 buildings from 72 universities. It includes 50,218 drone-view images for training and 37,855 for testing, paired with corresponding satellite orthophotos. This dataset is particularly challenging due to the inclusion of drone images captured at varying distances and angles.

SUES-200 focuses on multi-altitude scenarios, covering 200 disparate locations. It provides drone images captured at four distinct altitudes (150m, 200m, 250m, and 300m). This multi-level height design allows for a fine-grained assessment of the model's robustness against scale variations.

Following standard protocols, we employ Recall at K (R@K) and Average Precision (AP) as the primary metrics. R@K measures the percentage of query images for which the ground truth is retrieved within the top-K predictions, while AP evaluates the precision of the retrieval ranking.
\subsection{Implementation Details}
\vspace{-2pt}
We implement (MGS)$^2$-Net using the PyTorch framework on a single NVIDIA A800 GPU. During the training phase, the input images for both satellite and drone views are resized to $224 \times 224$. For inference, we adopt a higher resolution of $336 \times 336$ to capture fine-grained geometric details. We utilize a pre-trained DINOv2-B \cite{35} as the backbone to extract initial semantic features and AdamW with a weight decay of $9.5 \times 10^{-9}$. A linear learning rate schedule is applied. The batch size is 128. We employ parameter-wise learning rates: $2 \times 10^{-5}$ for the RGB backbone, $5 \times 10^{-5}$ for depth-related backbone (depth encoder + MGS-F + MGS-A), and $1 \times 10^{-4}$ for the aggregator. The hyper-parameters for SGC loss, i.e., $\lambda_{geo}$ and $\xi$, are empirically set to 1.0 and 0.5, respectively. During inference, we utilize the cosine similarity of the final fused embeddings for retrieval ranking.

To prevent overfitting and reduce computational overhead, we freeze the majority of the pre-trained backbone during the training phase. Consequently, our (MGS)$^2$-Net framework comprises a total of 115.56M parameters, of which only 32.91M are trainable. This parameter-efficient design allows the network to focus on optimizing the proposed MGS-A and MGS-F modules, ensuring fast convergence and low training costs.

\vspace{-2pt}
\subsection{Comparison with State-of-the-Art Methods}
\subsubsection{Performance on University-1652.}

\begin{table}[!t]
\centering
\caption{Performance Comparison on the University-1652 Dataset. The Best Results Are in Bold.}
\label{tab:university}
\setlength{\tabcolsep}{3.5pt} 
\begin{tabular}{l c cc cc} 
\toprule
\multirow{2}{*}{Method} & \multirow{2}{*}{Test Image Size} & \multicolumn{2}{c}{Drone $\rightarrow$ Satellite} & \multicolumn{2}{c}{Satellite $\rightarrow$ Drone} \\
\cmidrule(lr){3-4} \cmidrule(lr){5-6}
& & R@1 & AP & R@1 & AP \\
\midrule
Safe-Net \cite{51}         & 256 $\times$ 256 & 86.98 & 88.85 & 91.22 & 86.06 \\
GeoFormer \cite{37}    & 224 $\times$ 224 & 89.08 & 90.83 & 92.30 & 88.54 \\
MCCG \cite{44}         & 384 $\times$ 384 & 89.64 & 91.32 & 94.30 & 89.39 \\
SDPL \cite{45}         & 256 $\times$ 256 & 90.16 & 91.64 & 93.58 & 89.45 \\
CCR \cite{46}          & 384 $\times$ 384 & 92.54 & 93.78 & 95.15 & 91.80 \\
Sample4Geo \cite{7}    & 384 $\times$ 384 & 92.65 & 93.81 & 96.43 & 93.79 \\
SRLN \cite{23}         & 384 $\times$ 384 & 92.70 & 93.77 & 95.14 & 91.97 \\
MEAN \cite{24}         & 384 $\times$ 384 & 93.55 & 94.53 & 96.01 & 92.08 \\
SCOF \cite{47}         & 384 $\times$ 384 & 93.68 & 94.68 & 96.29 & 92.68 \\
CAMP \cite{8}          & 384 $\times$ 384 & 94.46 & 95.38 & 96.15 & 92.72 \\
DAC \cite{48}          & 384 $\times$ 384 & 94.67 & 95.50 & 96.43 & 93.79 \\
QDFL \cite{25}         & 280 $\times$ 280 & 95.00 & 95.83 & 97.15 & 94.57 \\
CDM-Net \cite{27}      & 512 $\times$ 512 & 95.13 & 96.04 & 96.43 & 93.79 \\
JRN-Geo \cite{50}      & 384 $\times$ 384 & 95.13 & 95.85 & 96.72 & 94.93 \\

\midrule
\textbf{Ours} & 336 $\times$ 336 & \textbf{97.60} & \textbf{98.03} & \textbf{98.86} & \textbf{97.25} \\
\bottomrule
\end{tabular}
\end{table}

We compare (MGS)$^2$-Net against recent state-of-the-art (SOTA) methods, including geometric-aware approaches like GeoFormer and semantic-enhanced methods like CAMP\cite{8} and QDFL\cite{25}. The quantitative results are summarized in Table \ref{tab:university}. 

It can be seen that (MGS)$^2$-Net achieves superior performance across all metrics. In particular, in the Drone$\rightarrow$Satellite task, our method achieves a Recall@1 of 97.60\% and an AP of 98.03\%. This represents a significant improvement over the runner-up JRN-Geo \cite{50} (+2.47\% in R@1) and QDFL \cite{25} (+2.60\% in R@1). Unlike methods that treat the image as a 2D plane, our framework explicitly models the 3D structure. The substantial gain indicates that our method ensures that hard-negative facade features do not confuse the retrieval ranking.
\vspace{-5pt}
\subsubsection{Robustness to Scale Variations on SUES-200.}

\begin{table}[!t]
\centering
\caption{Performance Comparison Between Different Methods on Drone-to-Satellite and Satellite-to-Drone Retrieval Tasks. The Best Results Are in Bold.}
\label{tab:sues200}
\scriptsize 
\renewcommand{\arraystretch}{1.12}
\setlength{\tabcolsep}{0pt} 
\begin{tabular*}{\textwidth}{@{\extracolsep{\fill}} l c cc cc cc cc} 
\toprule
\multicolumn{10}{c}{Drone$\rightarrow$Satellite} \\ 
\midrule
\multirow{2}{*}{Method} & \multirow{2}{*}{Test Image Size} & \multicolumn{2}{c}{150m} & \multicolumn{2}{c}{200m} & \multicolumn{2}{c}{250m} & \multicolumn{2}{c}{300m} \\
\cmidrule(lr){3-4} \cmidrule(lr){5-6} \cmidrule(lr){7-8} \cmidrule(lr){9-10} 
& & R@1 & AP & R@1 & AP & R@1 & AP & R@1 & AP \\
\midrule
SUES-200 \cite{39} & 384 $\times$ 384 & 59.32 & 64.93 & 62.30 & 67.24 & 71.35 & 75.49 & 77.17 & 67.80 \\
Safe-Net \cite{51} & 256 $\times$ 256 & 81.05 & 84.76 & 91.10 & 93.04 & 94.52 & 95.74 & 94.57 & 95.60 \\
MCCG \cite{44} & 384 $\times$ 384 & 82.22 & 85.47 & 89.38 & 91.41 & 93.82 & 95.04 & 95.07 & 96.20 \\
SDPL \cite{45} & 256 $\times$ 256 & 82.95 & 85.82 & 92.73 & 94.07 & 96.05 & 96.69 & 97.83 & 98.05 \\
CCR \cite{46} & 384 $\times$ 384 & 87.08 & 89.55 & 93.57 & 94.90 & 95.42 & 96.28 & 96.82 & 97.39 \\
Sample4Geo \cite{7} & 384 $\times$ 384 & 92.60 & 96.38 & 97.38 & 97.81 & 98.28 & 98.64 & 99.18 & 99.36 \\
SRLN \cite{23} & 384 $\times$ 384 & 89.90 & 91.90 & 94.32 & 95.65 & 95.92 & 96.79 & 96.37 & 97.21 \\
SCOF \cite{47} & 384 $\times$ 384 & 90.75 & 92.32 & 94.25 & 95.35 & 96.88 & 97.42 & 97.85 & 98.10 \\
QDFL \cite{25} & 280 $\times$ 280 & 93.97 & 95.42 & 98.25 & 98.67 & 99.30 & 99.48 & 99.31 & 99.48 \\
CDM-Net \cite{27} & 512 $\times$ 512 & 93.78 & 95.16 & 97.62 & 98.16 & 98.28 & 98.69 & 99.20 & 99.31 \\
CAMP \cite{8} & 384 $\times$ 384 & 95.40 & 96.38 & 97.63 & 98.16 & 98.05 & 98.45 & 99.33 & 99.46 \\
MEAN \cite{24} & 384 $\times$ 384 & 95.50 & 96.46 & 98.38 & 98.72 & 98.95 & 99.17 & 99.52 & 99.63 \\
DAC \cite{48} & 384 $\times$ 384 & 96.80 & 97.54 & 97.48 & 97.97 & 98.20 & 98.62 & 97.58 & 98.14 \\
\midrule
\textbf{Ours} & 336 $\times$ 336 & \textbf{98.45} & \textbf{98.78} & \textbf{99.62} & \textbf{99.69} & \textbf{99.78} & \textbf{99.80} & \textbf{100.00} & \textbf{100.00} \\
\toprule
\multicolumn{10}{c}{Satellite$\rightarrow$Drone} \\ 
\midrule
\multirow{2}{*}{Method} & \multirow{2}{*}{Test Image Size} & \multicolumn{2}{c}{150m} & \multicolumn{2}{c}{200m} & \multicolumn{2}{c}{250m} & \multicolumn{2}{c}{300m} \\
\cmidrule(lr){3-4} \cmidrule(lr){5-6} \cmidrule(lr){7-8} \cmidrule(lr){9-10}
& & R@1 & AP & R@1 & AP & R@1 & AP & R@1 & AP \\
\midrule
SUES-200 & 384 $\times$ 384 & 82.50 & 58.95 & 85.00 & 62.56 & 88.75 & 69.96 & 96.25 & 84.16 \\
MCCG & 384 $\times$ 384 & 93.75 & 89.72 & 93.75 & 92.21 & 96.25 & 96.14 & 98.75 & 96.64 \\
SDPL & 384 $\times$ 384 & 93.75 & 83.75 & 96.25 & 92.42 & 97.50 & 95.65 & 96.25 & 96.17 \\
CCR & 384 $\times$ 384 & 92.50 & 88.54 & 97.50 & 95.22 & 97.50 & 97.10 & 97.50 & 97.49 \\
Sample4Geo & 384 $\times$ 384 & 97.50 & 93.63 & 98.75 & 96.70 & 98.75 & 98.28 & 98.75 & 98.05 \\
SRLN & 384 $\times$ 384 & 93.75 & 93.01 & 97.50 & 95.08 & 97.50 & 96.52 & 97.50 & 96.71 \\
SCOF & 384 $\times$ 384 & 95.00 & 89.72 & 97.50 & 93.13 & 98.75 & 96.33 & 97.50 & 96.62 \\
CDM-Net & 384 $\times$ 384 & 95.25 & 92.24 & 98.50 & 96.40 & 99.00 & 97.60 & 99.00 & 98.01 \\
CAMP & 384 $\times$ 384 & 96.25 & 93.69 & 97.50 & 96.76 & 98.75 & 98.10 & \textbf{100.00} & 98.85 \\
Safe-Net & 256 $\times$ 256 & 97.50 & 86.36 & 96.25 & 92.61 & 97.50 & 94.98 & 98.75 & 95.67 \\
DAC & 384 $\times$ 384 & 97.50 & 94.06 & 98.75 & 96.66 & 98.75 & 98.09 & 98.75 & 97.87 \\
MEAN & 384 $\times$ 384 & 97.50 & 94.75 & \textbf{100.00} & 97.09 & \textbf{100.00} & 98.28 & \textbf{100.00} & \textbf{99.21} \\
QDFL & 384 $\times$ 384 & \textbf{98.75} & 95.10 & 98.75 & 97.92 & \textbf{100.00} & \textbf{99.07} & \textbf{100.00} & 99.07 \\
\midrule
\textbf{Ours} & 336 $\times$ 336 & \textbf{98.75} & \textbf{96.50} & \textbf{100.00} & \textbf{98.51} & \textbf{100.00} & 98.73 & \textbf{100.00} & 98.95 \\
\bottomrule
\end{tabular*}
\end{table}

We also evaluate our method on SUES-200 across different flight altitudes. As shown in Table \ref{tab:sues200}, (MGS)$^2$-Net exhibits remarkable robustness. While most existing methods, such as CDM-Net \cite{27}, suffer from performance degradation at lower altitudes (150m) due to the impact of more vertical facades, (MGS)$^2$-Net maintains a high R@1 of 98.45\% at 150m on the Drone$\rightarrow$Satellite task. As the altitude increases to 300m, the performance of the proposed method reaches the maximum value, achieving a perfect 100\% R@1. The reason for maintaining extremely high recall at high altitudes is that our MGS-A module provides a more physically grounded solution to scale ambiguity. Compared to MEAN \cite{24}, which employs multi-level embeddings, our approach still outperforms it at 300m altitude. Notably, on the Satellite$\rightarrow$Drone task, apart from the challenging 150m altitude, the results on other altitudes have achieved 100\% R@1. This indicates that the localization task in this scenario has been completely resolved.
\vspace{-5pt}
\subsubsection{Cross-Dataset Generalization.}

A critical bottleneck for learning-based CVGL is overfitting to the source domain's specific architectural style. To comprehensively test generalization capabilities, we train our model solely on the University-1652 dataset and directly evaluate it under two distinct zero-shot transfer scenarios: varying altitudes (SUES-200) and severe stylistic shifts (DenseUAV). 

\begin{figure}[!tp]
    \centering
    \includegraphics[width=12cm]{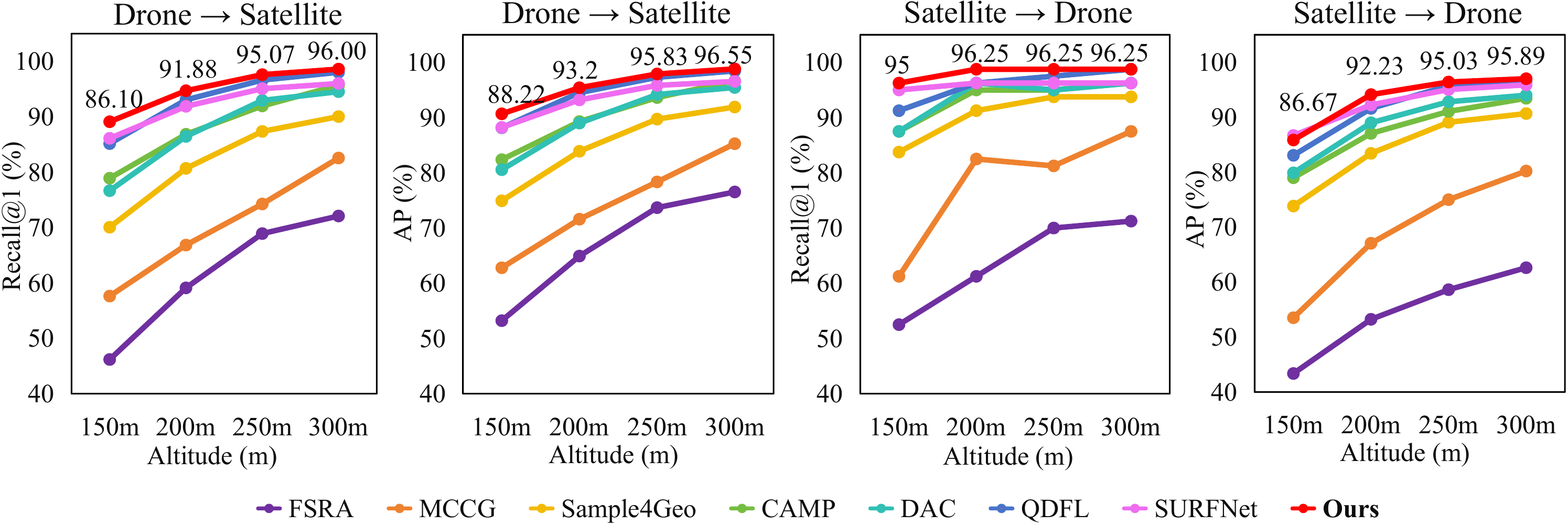}
    \caption{Visualizations of generalization results on SUES-200 dataset compared with state-of-the-art methods, using University-1652 only as the  training set.}
    \label{fig5}
\end{figure}

First, we evaluate the generalization performance on the SUES-200 dataset to verify robustness against unseen scale variations. As illustrated in Fig. \ref{fig5}, despite the absence of target-domain training data, (MGS)$^2$-Net consistently achieves the best retrieval performance across all altitude sub-datasets when compared with state-of-the-art methods. This demonstrates that our scale adaptation mechanism effectively generalizes to novel environments.

To further test the model's robustness, we evaluate it on the DenseUAV dataset \cite{53}, as detailed in Table \ref{cross}. In this extreme zero-shot transfer case, most existing methods collapse on the retrieval task. As noted by SURFNet \cite{10}, DenseUAV presents a significant challenge because it possesses stronger photometric and stylistic domain gaps, alongside much denser urban coverage. Consequently, methods relying on texture or passive attention fail to generalize. For instance, CAMP \cite{8} yields only 23.48\% R@1. In contrast, our method maintains a remarkable Recall@1 of 84.60\%. 

\begin{table}[!t]
\centering
\caption{Generalization from University‑1652 to DenseUAV.}
\label{cross}
\small
\setlength{\tabcolsep}{4pt}
\begin{tabular}{l cc cc}
\toprule
Method & \multicolumn{2}{c}{UAV$\rightarrow$Satellite} & \multicolumn{2}{c}{Satellite$\rightarrow$UAV} \\
\cmidrule(lr){2-3} \cmidrule(lr){4-5}
& R@1 & AP & R@1 & AP \\
\midrule
Sample4Geo \cite{7} & 22.00 & 14.58 & 19.10 & 17.46 \\
CAMP \cite{8} & 23.48 & 15.75 & 20.97 & 20.29 \\
SURFNet (Zero shot) \cite{10} & 24.28 & 16.13 & 21.04 & 20.37 \\
SURFNet (Fine-tune) \cite{10} & 71.06 & 65.99 & \textbf{87.43} & 51.29 \\
\midrule
\textbf{Ours (Zero shot)} & \textbf{84.60} & \textbf{67.33} & 77.05 & \textbf{68.89} \\
\bottomrule
\end{tabular}
\end{table}

This dramatic performance gap suggests that (MGS)$^2$-Net learns intrinsic geometric properties rather than dataset-specific texture patterns. By physically filtering out non-transferable vertical facades, the model robustly focuses on the structural "fingerprint" of co-visible planes, which inherently remains invariant across diverse cities and altitudes.

\vspace{-5pt}
\subsection{Ablation Studies}
\vspace{-3pt}
To verify the individual contributions of each component in (MGS)$^2$-Net, we conducted a comprehensive step-wise ablation study on the University-1652 drone-to-satellite retrieval task. The quantitative results are summarized in Table \ref{tab:ablation}.

\begin{table}[!t]
\centering
\caption{Component-wise Ablation Analysis on the University‑1652 Drone-to-Satellite Retrieval Task.}
\label{tab:ablation}
\small
\setlength{\tabcolsep}{4pt}
\begin{tabular}{lccccc}
\toprule
Model & MGS-A & MGS-F & SGC & R@1 & AP \\
\midrule
Baseline I (w/o Depth) &  &  &  & 92.05 & 93.34 \\
Baseline II (w/ Depth) &  &  &  & 94.24 & 95.02 \\
\midrule
Ours I &  $\checkmark$  &  &  & 95.13$_{+0.89}$ & 95.90$_{+0.88}$\\
Ours II & $\checkmark$ & $\checkmark$ &  & 97.32$_{+3.08}$ & 97.67$_{+2.65}$ \\
\textbf{Ours III (Full)} & $\checkmark$ & $\checkmark$ & $\checkmark$ & \textbf{97.60}$_\mathbf{+3.36}$ & \textbf{98.03}$_\mathbf{+3.01}$ \\
\bottomrule
\end{tabular}
\end{table}

Starting from the Baseline I (DINOv2-B \cite{35} + SALAD \cite{54}, without depth prior), which yields 92.05\% Recall@1, we first validate the necessity of the depth modality and its integration strategy. To further show the architectural contribution, Baseline II implements a naive fusion baseline by explicitly concatenating the depth features with Baseline I. While this simple concatenation provides a noticeable baseline improvement (to 94.24\% R@1), it remains sub-optimal for handling the severe scale ambiguity caused by varying drone flight altitudes. The improvement indicated by the subscript numbers in the lower half of Table \ref{tab:ablation} is measured relative to Baseline II.

By replacing the naive concatenation with our MGS-A module (Ours I), the network explicitly utilizes depth priors to dynamically re-weight multi-scale features. While this yields a modest direct numerical gain (to 95.13\% R@1), its primary architectural contribution lies in functionally aligning the multi-scale features with physical depth, serving as a critical geometric foundation for subsequent processing.

Building upon this, the addition of the MGS-F module (Ours II) further elevates the R@1 to 97.32\% (a +2.19\% increase). A qualitative analysis and visualization of the MGS-F module, along with further discussion of its mechanism of action, will be presented in the next subsection.

Finally, combining both modules with the SGC Loss (Ours III) yields the full (MGS)$^2$-Net framework, achieving the best performance. The SGC Loss acts as a crucial rectifier during the training phase, ensuring that the feature representations inherently maintain the correct geometric activation ordering even when facing complex architectural layouts.

\vspace{-5pt}
\subsection{Qualitative Analysis and Visualization}
\vspace{-3pt}
To provide intuitive insights into how (MGS)$^2$-Net bridges the domain gap, we present visualizations of feature responses and retrieval results.

\begin{figure}[!t]
    \centering
    \includegraphics[width=12cm]{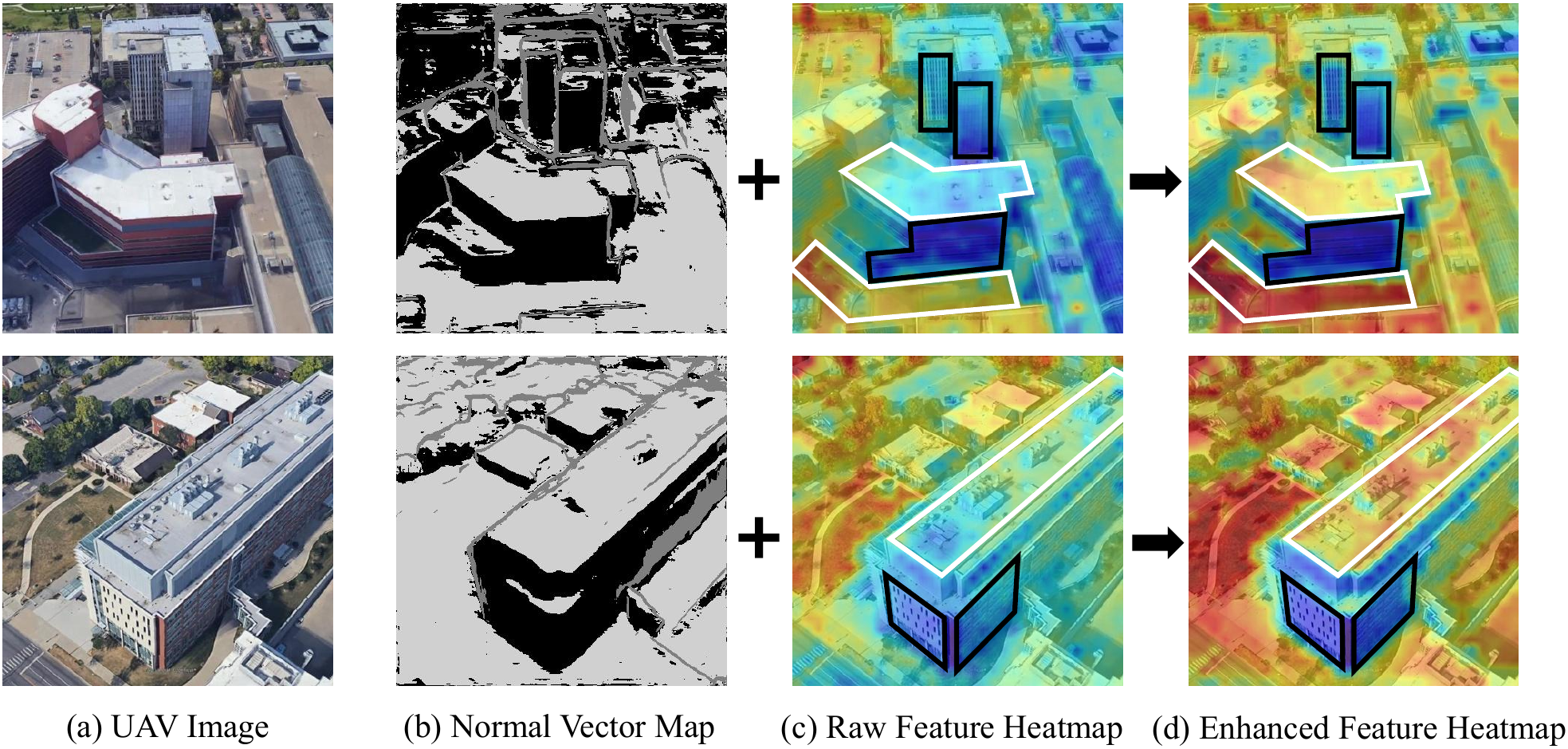}
    \caption{Visualization of the MGS-F mechanism. We visualize the feature response maps before and after applying MGS-F. (b) The white area indicates that the normal vector points vertically upward, while the black area indicates that the normal vector is nearly parallel to the ground. (c) The backbone naively activates texture-rich regions. (d) MGS-F redistributes attention based on geometric priors. Note that black boxes indicate the effective suppression of view-dependent vertical facades, while white boxes highlight the adaptive enhancement of view-invariant horizontal planes.}
    \label{fig3}
\end{figure}

\vspace{-3pt}
\subsubsection{Visualization of Geometric Filtering.}
Fig. \ref{fig3} visualizes the feature activation maps before and after applying MGS-F. Fig. \ref{fig3}(b) clearly demonstrates that MGS-F successfully predicts horizontal planes (white area) and vertical planes (black area). Using this as a mask, it forces the feature map to focus more on horizontal plane features rather than vertical ones. This is also well indicated on the feature map: horizontal planes within the white box are effectively enhanced, while vertical planes within the black box are suppressed. This verifies that MGS-F functions not merely as an attention mechanism, but as a physical filter that aligns the drone view with the orthographic satellite map.

\begin{figure}[!tb]
    \centering
    \includegraphics[width=10cm]{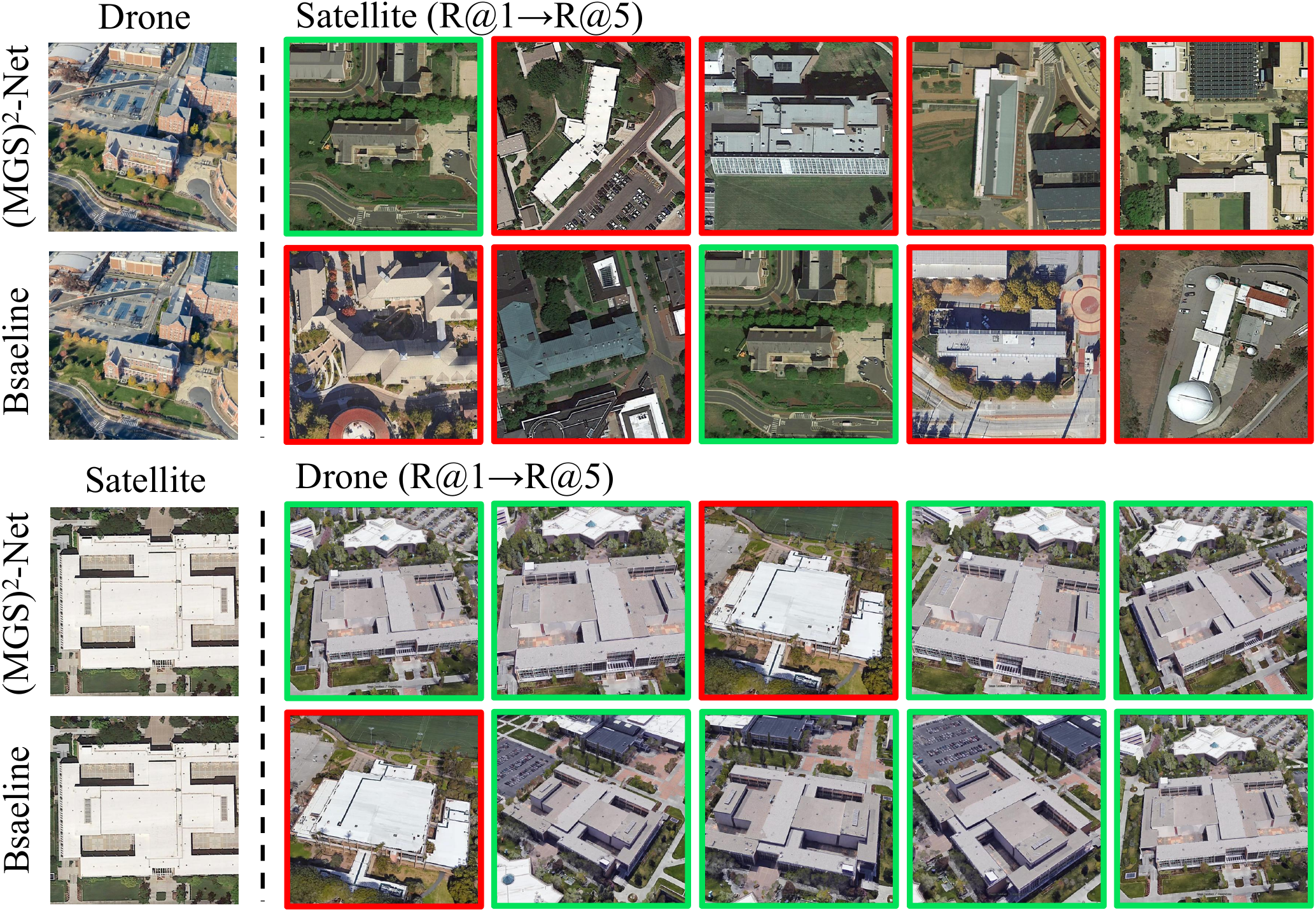}
    \caption{Qualitative retrieval results for the University-1652 dataset. (Top) Top 5 retrieval results for the drone to satellite retrieval task.  (Bottom) Top 5 retrieval results for the satellite to drone retrieval task. The first and third rows show retrieval results for (MGS)$^2$-Net, while the second and fourth rows show retrieval results for the baseline. \textcolor{ForestGreen}{Green} borders indicate correct matches and \textcolor{red}{red} borders indicate incorrect matches.}
    \label{fig4}
\end{figure}
\vspace{-3pt}
\subsubsection{Retrieval Results.}
We further compared the top 5 retrieval results between the baseline and (MGS)$^2$-Net and selected representative examples as shown in Figure \ref{fig4}. As demonstrated in the second and fourth rows, the baseline retrieves candidates with similar facade textures but different structural layouts, leading to false positives. For instance, in the UAV$\rightarrow$Satellite task, the baseline (row 2) tends to recall results containing red features during retrieval because the query facade texture exhibits red features. In contrast, (MGS)$^2$-Net (row 1) correctly retrieves the target location while ignoring the red features. In the Satellite$\rightarrow$UAV task, a challenging case involves a roof with large white features. (MGS)$^2$-Net actively focuses on the roof structure while suppressing excessive RGB influence, finding the correct candidate. This is particularly common in densely built environments, where our method's ability to suppress non-co-visible regions in cross-view pairs is crucial for accurate localization.

\vspace{-5pt}
\section{CONCLUSION\label{5} }
\vspace{-5pt}
In this paper, we presented (MGS)$^2$-Net, a geometry-grounded framework that bridges the UAV-satellite domain gap by shifting from 2D appearance matching to 3D structural alignment. By synergizing the Macro-Geometric Structure Filtering (MGS-F) to suppress vertical interferences and the Micro-Geometric Scale Adaptation (MGS-A) to rectify scale discrepancies, our method effectively extracts view-invariant geometric features. Extensive experiments demonstrate that (MGS)$^2$-Net achieves state-of-the-art performance on University-1652 and SUES-200 benchmarks. Future work will focus on distilling the geometric awareness into a lightweight backbone for real-time edge deployment and improving robustness against extreme illumination variations.






\bibliographystyle{unsrt}
\bibliography{reference}
\end{document}